\def\year{2020}\relax
\documentclass[letterpaper]{article} 
\usepackage{aaai20}  
\usepackage{times}  
\usepackage{helvet} 
\usepackage{courier}  
\usepackage[hyphens]{url}  
\usepackage{graphicx} 
\usepackage{graphicx}  
\nocopyright
\title{Generate (non-software) Bugs to Fool Classifiers}
\author{Hiromu Yakura\textsuperscript{\rm 1,2}\thanks{Also with Teambox Inc., Japan.},
Youhei Akimoto\textsuperscript{\rm 1,2}, Jun Sakuma\textsuperscript{\rm 1,2} \\
\textsuperscript{\rm 1} University of Tsukuba, Japan \\
\textsuperscript{\rm 2} RIKEN Center for Advanced Intelligence Project, Japan \\
hiromu@mdl.cs.tsukuba.ac.jp, \{akimoto, jun\}@cs.tsukuba.ac.jp
}

\usepackage{algorithm}
\usepackage{algorithmic}
\usepackage{amsfonts}
\usepackage{amsmath}
\usepackage{bm}
\usepackage{enumitem}

\newcommand{\citet}[1]{\citeauthor{#1} \shortcite{#1}}
\newcommand{\citep}{\cite}
\newcommand{\citealp}[1]{\citeauthor{#1} (\citeyear{#1})}

\newcommand{\secref}[1]{\mbox{Section~\ref{#1}}}
\newcommand{\figref}[1]{\mbox{Figure~\ref{#1}}}

\newcommand{\eqnref}[1]{\mbox{Equation~\eqref{#1}}}
\newcommand{\algref}[1]{\mbox{Algorithm~\ref{#1}}}

\begin{document}

\maketitle

\begin{abstract}
In adversarial attacks intended to confound deep learning models, most studies have focused on limiting the magnitude of the modification so that humans do not notice the attack.
On the other hand, during an attack against autonomous cars, for example, most drivers would not find it strange if a small insect image were placed on a stop sign, or they may overlook it.
In this paper, we present a systematic approach to generate natural adversarial examples against classification models by employing such natural-appearing perturbations that imitate a certain object or signal.
We first show the feasibility of this approach in an attack against an image classifier by employing generative adversarial networks that produce image patches that have the appearance of a natural object to fool the target model.
We also introduce an algorithm to optimize placement of the perturbation in accordance with the input image, which makes the generation of adversarial examples fast and likely to succeed.
Moreover, we experimentally show that the proposed approach can be extended to the audio domain, for example, to generate perturbations that sound like the chirping of birds to fool a speech classifier.
\end{abstract}

\section{Introduction}
\label{sec:introduction}

Despite the great success of deep learning in various fields \cite{DBLP:journals/nature/LeCunBH15}, recent studies have shown that deep learning methods are vulnerable to adversarial examples \cite{DBLP:journals/corr/SzegedyZSBEGF13,DBLP:journals/corr/GoodfellowSS14}.
In other words, an attacker can make deep learning models misclassify examples by intentionally adding small perturbations to the examples, which are referred to as adversarial examples.
Following a study by \citealp{DBLP:journals/corr/SzegedyZSBEGF13} on image classification, adversarial examples have been demonstrated in many other domains,
including natural language processing \cite{DBLP:conf/emnlp/JiaL17}, speech recognition \cite{DBLP:conf/sp/Carlini018}, and malware detection \cite{DBLP:conf/esorics/GrossePMBM17}. 
Moreover, some studies \cite{DBLP:conf/cvpr/EykholtEF0RXPKS18,DBLP:conf/pkdd/ChenCMC18} have demonstrated a practical attack scenario based on adversarial examples to make autonomous driving systems misclassify stop signs by placing stickers on them.

These adversarial examples are most commonly generated by perturbing the input data, in a way that limits the magnitude of the perturbation so that humans do not notice the difference between a legitimate input sample and an adversarial example.
When image classification models are attacked, regularization by $L_2$- or $L_\infty$-norm is often used to make the generated adversarial examples unnoticeable to humans.
In another approach, some studies \cite{DBLP:conf/ijcai/XiaoLZHLS18,DBLP:conf/iclr/ZhaoDS18} introduced generative adversarial networks (GAN) \cite{DBLP:conf/nips/GoodfellowPMXWOCB14} to prepare adversarial examples that are likely to appear as natural images.
In these methods, GAN is used to generate adversarial examples that are close to the distribution of pre-given natural images.

\begin{figure}[t]
    \centering
    \includegraphics[width=0.4\textwidth]{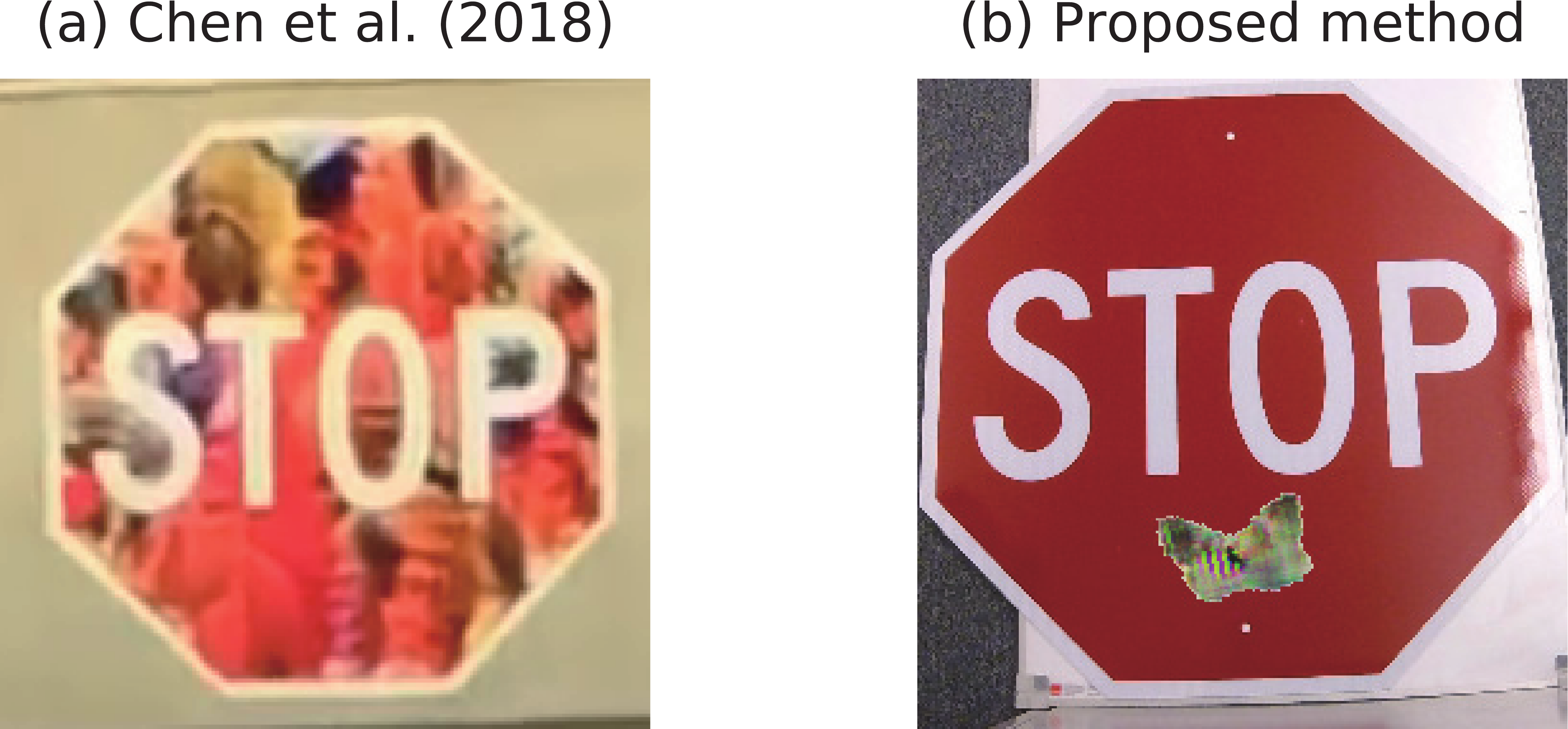}
    \caption{Because road signs are commonly located outside, adding a bug image to such a sign is more natural than perturbing with an arbitrary pattern and is more difficult to notice.}
    \label{fig:overview}
\end{figure}

One drawback of the GAN-based approach is that natural adversarial examples do not always necessarily resemble the input or natural images in some attack scenarios.
For example, to attack autonomous cars, placing small bugs on a stop sign would be more natural than modifying the entire sign, as shown in \figref{fig:overview}.
More generally, by overlaying a perturbation that imitates a natural object consistent with the attack scenario on part of the input image, we can create adversarial examples less likely to be noticed.
Such a strategy can be extended to domains other than images, for example, to attack speech classification models by generating a perturbation that sounds like environmental noise.

Given the widespread use of image classification and speech recognition in real-world applications to replace human involvement, it is important to analyze such gaps between these models and human perception.
In particular, if we can investigate them systematically, it would be possible to design models and training processes to overcome such gaps automatically.

In this paper, we propose a systematic approach to generate natural adversarial examples against classification models by making perturbations that mimic objects or sound signals that are unnoticeable.
In this method, we use GAN to generate perturbations so that they imitate collections of reference objects or signals while fooling the target model.
Here, because the attack against an image classifier uses a perturbation that is small in relation to the overall image,
it is expected that the classification results from the model would be affected not only by the content of the perturbation but also by its location in the image.
In other words, optimizing the location can increase the success rate of the attack, but it has the problem that the gradient at the location easily vanishes.
Thus, we also introduce a policy gradient method \cite{DBLP:journals/nn/SehnkeORGPS10}, which is typically used for reinforcement learning, to optimize the location without a gradient.

To confirm the effectiveness of the proposed method, we first conducted experiments with a road sign classifier and an ImageNet classifier.
The results confirmed that the obtained adversarial examples could fool an ImageNet classifier, even with bug image perturbations measuring $32\times32$ pixels, by optimizing the placement location.
Moreover, we demonstrate that the proposed approach can also be applied to a speech command classifier with a perturbation that sounds like the chirping of birds.
This paper presents a new strategy for creating unnoticeable adversarial attacks that manipulate the content of the perturbation to match the attack scenario.

\section{Background}
\label{sec:background}

As explained in \secref{sec:introduction}, most approaches use optimization algorithms to generate adversarial examples.
Let us consider a scenario in which an attacker wishes to modify input image $\bm{x}$ so that the target model $f$ classifies it with the specific label $t$.
The generation process can be represented as follows:
\begin{equation}
    \hat{\bm{v}} = {\rm argmin}_{\bm{v}} \ \mathcal{L}_{f} \left( \bm{x} + \bm{v}, t \right) + \epsilon \left \lVert \bm{v} \right \rVert \enspace,
    \label{eq:ae-vanilla}
\end{equation}
where $\mathcal{L}_{f}$ denotes a loss function that represents how distant the input data are from the given label under $f$
and $\bm{v} \mapsto \left \lVert \bm{v} \right \rVert$ is a norm function to regularize the perturbation so that $\bm{v}$ becomes unnoticeable to humans.
Then, $\bm{x} + \hat{\bm{v}}$ is expected to form an adversarial example that is classified as $t$ while it looks similar to $\bm{x}$.

Earlier approaches, such as \citealp{DBLP:journals/corr/SzegedyZSBEGF13} and \citealp{DBLP:conf/cvpr/Moosavi-Dezfooli16}, used $L_2$-norm to limit the magnitude of the perturbation.
In contrast, \citealp{DBLP:journals/corr/abs-1710-08864} used $L_0$-norm to limit the number of modified pixels and showed that even modification of a one-pixel could generate adversarial examples.

More recent studies introduced GAN instead of directly optimizing perturbations \cite{DBLP:conf/ijcai/XiaoLZHLS18,DBLP:conf/iclr/ZhaoDS18} for the purpose of ensuring the naturalness of adversarial examples.
For example, \citealp{DBLP:conf/ijcai/XiaoLZHLS18} trained a discriminator network to distinguish adversarial examples from natural images so that the generator network produced adversarial examples that appeared as natural images.
Given the distribution $p_{\bm{x}}$ over the natural images and the trade-off parameter $\alpha$, its training process can be represented similarly to that in \citealp{DBLP:conf/nips/GoodfellowPMXWOCB14} as follows:
\begin{eqnarray}
    & \min_{\mathcal{G}} \max_{\mathcal{D}} \! \! \! \! & \mathbb{E}_{\bm{x} \sim p_{\bm{x}}} \left[ \log \mathcal{D} \left( \bm{x} \right) \right] \nonumber \\
    &                                                   & + \, \mathbb{E}_{\bm{x} \sim p_{\bm{x}}} \left[ \log \left( 1 - \mathcal{D} \left( \bm{x} + \mathcal{G} \left( \bm{x} \right) \right) \right) \right] \nonumber \\
    &                                                   & + \, \alpha \, \mathbb{E}_{\bm{x} \sim p_{\bm{x}}} \left[ \mathcal{L}_{f} \left( \bm{x} + \mathcal{G} \left( \bm{x} \right) \! , t \right) \right] \enspace.
    \label{eq:advgan}
\end{eqnarray}
Then, we can obtain the generator network $\mathcal{G}$ that outputs a perturbation over the entire region of the input image so that the overlaid image fools the target model.

However, we want to generate adversarial examples by placing small objects in the image, not by modifying the entire image.
In that respect, \citealp{DBLP:journals/corr/abs-1712-09665} proposed an adversarial patch, which changes the classification results by placing the patch at an arbitrary location in the input image.
Given that $A \left( \bm{p}, \bm{x}, \bm{\theta} \right)$ represents an operation of placing a patch $\bm{p}$ in image $\bm{x}$ with the location and rotation specified by the parameter $\bm{\theta}$, its generation process is represented as follows:
\begin{eqnarray}
    \hat{\bm{p}} = {\rm argmin}_{\bm{p}} \ \mathbb{E}_{\bm{\theta} \sim p_{\bm{\theta}}} \left[ \mathcal{L}_{f} \left( A \left( \bm{p}, \bm{x}, \bm{\theta} \right) \! , t \right) \right] \enspace,
    \label{eq:adv-patch}
\end{eqnarray}
where $p_{\bm{\theta}}$ is the distribution of all possible locations and rotations.
We note that this method generates patches with ignoring the attack context, and thus the resulting patches appear to be arresting artifacts that would be instantly recognized by humans.

In contrast, \citealp{DBLP:conf/ccs/SharifBBR16} proposed an attack against a face recognition model by generating perturbations within a fixed frame that look like eyeglasses.
As a result, the person in their adversarial example appears to wear glasses with a strange pattern.
Though their approach for making the attack inconspicuous is quite similar to our idea, they only focused on the shape of the perturbation, not its content.
Thus, the application of their method is limited to cases in which a strange pattern is acceptable, such as using eccentrically colored eyeglasses to attack a facial recognition model.
In addition, this method requires human attackers to design the shape and location of the frame specifically and is not applicable to the audio domain.

Based on the articles reviewed above, we concluded that existing methods minimize the perceptibility by humans mainly by focusing on limited aspects of a perturbation, such as its magnitude or size.
In contrast, our approach focuses on the content of perturbations themselves and makes them hard to notice by imitating a natural object or signal, such as bug images or bird songs, to disguise the deception.
In addition, unlike \citealp{DBLP:conf/ccs/SharifBBR16}, the proposed method can be systematically used in a wide range of attack scenarios because the GAN can mimic arbitrary objects.
We believe that a new strategy of unnoticeable attacks can be created by manipulating the content of the perturbations to conform with the attack situation.

\section{Proposed Method}
\label{sec:method}

In this paper, we propose a method to generate adversarial examples by placing small perturbations that look similar to a familiar object, such as bugs.
We propose two approaches to cover a wide range of attack scenarios.
The first one is a patch-based method, which makes the obtained adversarial examples robust against different placement locations.
The second one is based on the policy gradient method, which is designed to increase the success rate instead of preserving the robustness of the location change.

\subsection{Patch-based Method}
\label{sec:method-patch}

As described in \secref{sec:background}, \citealp{DBLP:conf/ijcai/XiaoLZHLS18} presented a method for making the adversarial examples similar to natural images of any class.
Our method extends their approach, shown in \eqnref{eq:advgan}, so that the obtained perturbations become similar to certain reference images (e.g., bug images).

At the same time, our method has a degree of freedom with regard to the location to place perturbations, unlike that embodied in \eqnref{eq:advgan}, because our aim is not to modify the entire region but to overlay small patches.
Thus, inspired by \citealp{DBLP:journals/corr/abs-1712-09665}, we introduce a mechanism to make the perturbations robust against changes in location for the same manner as that in \eqnref{eq:adv-patch}.
In other words, given that $p_{\bm{v}}$ represents a distribution over the reference images of the specific object, the proposed method is presented as follows:
\begin{eqnarray}
    & \! \! \! \! \! \! \min\limits_{\mathcal{G}} \max\limits_{\mathcal{D}} \! \! \! \! \! & \mathbb{E}_{\bm{v} \sim p_{\bm{v}}} \left[ \log \mathcal{D} \left( \bm{v} \right) \right] + \mathbb{E}_{\bm{z} \sim p_{\bm{z}}} \left[ \log \left( 1 \! - \! \mathcal{D} \left( \mathcal{G} \left( \bm{z} \right) \right) \right) \right] \nonumber \\
    &                                                                                      & + \, \alpha \, \mathbb{E}_{\bm{z} \sim p_{\bm{z}}, \bm{\theta} \sim p_{\bm{\theta}}} \left[ \mathcal{L}_{f} \left( A \left( \mathcal{G} \left( \bm{z} \right) \!, \bm{x}, \bm{\theta} \right) \! , t \right) \right] \enspace,
    \label{eq:proposed-patch}
\end{eqnarray}
where $p_{\bm{z}}$ represents the prior distribution for the generator network in the same manner used in \citealp{DBLP:conf/nips/GoodfellowPMXWOCB14}.

\subsection{PEPG-based Method}
\label{sec:method-pepg}

\secref{sec:method-patch} introduced \eqnref{eq:proposed-patch} to produce a perturbation that works without regard to the location in which it is placed.
However, compared to \eqnref{eq:adv-patch}, its objective variable is changed from the patch itself to the parameter of the network generating the patch.
Considering such complexity, the generation process of adversarial examples by the patch-based method is expected to be much harder to deal with.

Therefore, we then optimize the location of perturbations using the policy gradient method, as mentioned in \secref{sec:introduction}.
This is based on the observation that image classification models often leverage local information in the input image \cite{DBLP:conf/cvpr/NgYD15}.
In other words, it suggests that an optimal perturbation location exists for each input image to fool the target model.
If we can find such a location, the generation process would become easier.

One potential approach is to use a subdifferentiable affine transformation \cite{DBLP:conf/nips/JaderbergSZK15}.
In other words, we optimize the parameter of the affine transform to apply it to the perturbation instead of optimizing the location directly.
However, this idea is not applicable to our situation because the gradient of the parameter is almost zero.
This can be understood by the fact that a very small change in the parameter places the perturbation in the same location.
That is, the output probability from the target model is not affected by a small change, consequently, producing zero gradient for the parameter.

Therefore, we use a parameter-exploring policy gradient (PEPG) method \cite{DBLP:journals/nn/SehnkeORGPS10} to optimize the location and rotation of the perturbation.
This method assumes the distribution over the parameters and trains the hyperparameters of the distribution instead of the parameters themselves.
The advantage of the method is that it can explore a wide range of parameters in addition to not needing the gradients of the parameters.

When using the PEPG method, our method initially applies various locations sampled from the prior Gaussian distribution.
Then, based on the loss value from the target model during the trials, it gradually updates the hyperparameters so that the locations with small losses are likely to be sampled from the distribution.
By doing so, we can train the hyperparameters and the generator network simultaneously.

The detailed process is shown in \algref{alg:pepg}.
Here, we used symmetric sampling and reward normalization in the PEPG to accelerate convergence, as suggested by \citealp{DBLP:journals/nn/SehnkeORGPS10}.

\begin{algorithm}[t]
    \caption{PEPG-based adversarial example generation}
    \label{alg:pepg}
    \begin{algorithmic}
       \STATE {\bfseries Hyperparameters:} the batch size $m$, the importance of the loss from the target model $\alpha$, the step size of PEPG $\beta$, and the initial value of the distribution in PEPG $\bm{\mu}_\text{init}, \bm{\sigma}_\text{init}$
       \STATE \vspace{0.4em} Initialize $\bm{\mu} = \bm{\mu}_\text{init}, \bm{\sigma} = \bm{\sigma}_\text{init}$
       \REPEAT
          \STATE
          \begin{itemize}[leftmargin=0.8em, itemsep=2pt, parsep=0pt]
              \item[//\hspace{-1.4em}] \vspace{0.2em} \hspace{0.2em} Training of the discriminator
              \item[\textbullet\hspace{-1.2em}]                      Sample $m$ noises $ \{ \bm{z}^{(1)}, \ldots, \bm{z}^{(m)} \} $ from $ p_{\bm{z}} $
              \item[\textbullet\hspace{-1.2em}]                      Sample $m$ examples $ \{ \bm{v}^{(1)}, \ldots, \bm{v}^{(m)} \} $ from $ p_{\bm{v}} $
              \item[\textbullet\hspace{-1.2em}]                      Update the parameter of the discriminator using the following gradient: \\[0.1em]
                  \hspace{0.3em}                                     $ \nabla \frac{1}{m} \sum\limits_{i = 1}^{m} \left[ \log \mathcal{D} \left( \bm{v}^{(i)} \right) + \log \left( 1 - \mathcal{D} \left( \mathcal{G} \left( \bm{z}^{(i)} \right) \right) \right) \right] $
              \item[//\hspace{-1.4em}] \vspace{0.4em} \hspace{0.2em} Generation of adversarial examples
              \item[\textbullet\hspace{-1.2em}]                      Sample $m$ noises $ \{ \bm{z}^{(1)}, \ldots, \bm{z}^{(m)} \} $ from $ p_{\bm{z}} $
              \item[\textbullet\hspace{-1.2em}]                      Sample $m$ location and rotation parameters $ \{ \bm{\theta}^{(1)}, \ldots, \bm{\theta}^{(m)} \} $ from $ \mathcal{N} \left( \bm{\mu}, \bm{\sigma} \right) $
              \item[\textbullet\hspace{-1.2em}]                      Create adversarial examples for $i = 1, \ldots, m$: \\[0.1em]
                  \hspace{0.3em}                                     $ \tilde{\bm{x}}^{(i)} = \left( A \left( \mathcal{G} \left( \bm{z}^{(i)} \right), \bm{x}, \bm{\theta}^{(i)} \right) \right) $
              \item[//\hspace{-1.4em}] \vspace{0.4em} \hspace{0.2em} Training of the generator
              \item[\textbullet\hspace{-1.2em}]                      Calculate loss value for the adversarial examples $ \ell^{(i)} = \mathcal{L}_f \left( \tilde{\bm{x}}^{(i)}, t \right) $ for $i = 1, \ldots, m$
              \item[\textbullet\hspace{-1.2em}]                      Update the parameter of the discriminator using the following gradient: \\[0.1em]
                  \hspace{0.3em}                                     $ \nabla \frac{1}{m} \sum\limits_{i = 1}^{m} \left[ \log \left( 1 - \mathcal{D} \left( \mathcal{G} \left( \bm{z}^{(i)} \right) \right) \right) + \alpha \ell^{(i)} \right] $
              \item[//\hspace{-1.4em}] \vspace{0.4em} \hspace{0.2em} Training of the PEPG distribution
              \item[\textbullet\hspace{-1.2em}]                      Update $ \bm{\mu} $ and $ \bm{\sigma} $ using $ \{ -\ell^{(1)}, \ldots, -\ell^{(m)} \} $ as the rewards and $ \frac{1}{m} \sum_{i = 1}^{m} -\ell^{(i)} $ as the baseline reward based on the PEPG algorithm
          \end{itemize} \vspace{0.4em}
       \UNTIL{a sufficient number of adversarial examples recognized as $t$ are generated}
    \end{algorithmic}
\end{algorithm}

\section{Experimental Results}
\label{sec:results}

To test our approaches, we first performed a preliminary experiment with a road sign classifier to investigate the feasibility of the proposed methods.
Then, we conducted an experiment with an ImageNet classifier to confirm the availability of the methods against a wide range of input images and target classes.
In both experiments, we compared the patch-based and PEPG-based methods\footnote{The source code for both experiments is available at \url{https://github.com/hiromu/adversarial_examples_with_bugs}.}.

\begin{figure}[t]
    \centering
    \includegraphics[width=0.48\textwidth]{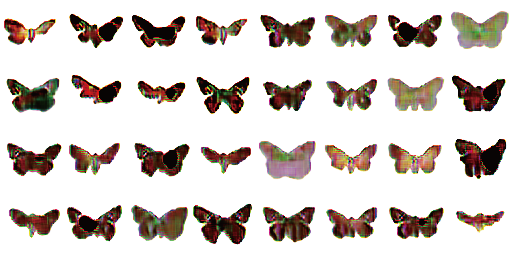}
    \caption{Examples of moth images generated by the pretrained GAN.}
    \label{fig:gan-moths}
\end{figure}

For the architecture of the GAN, we used WGAN-GP \cite{DBLP:conf/nips/GulrajaniAADC17} and changed the size of the output perturbations among $32 \times 32$, $64 \times 64$, and $128 \times 128$ pixels to evaluate the effect of the perturbation size.
For all tests, we used an image dataset of moths from Costa Rica \cite{DBLP:journals/corr/RodnerSBPWD15} for the reference images, that is, $p_{\bm{v}}$ in \eqnref{eq:proposed-patch}, to make the perturbations look like moths.
To ensure the stability of the training, we pretrained the GAN without the loss value from the target model and confirmed that it outputs moth-like images, as shown in \figref{fig:gan-moths}.

\subsection{Road Sign Classifier}
\label{sec:results-sign}

\begin{figure}[t]
    \centering
    \includegraphics[width=0.48\textwidth]{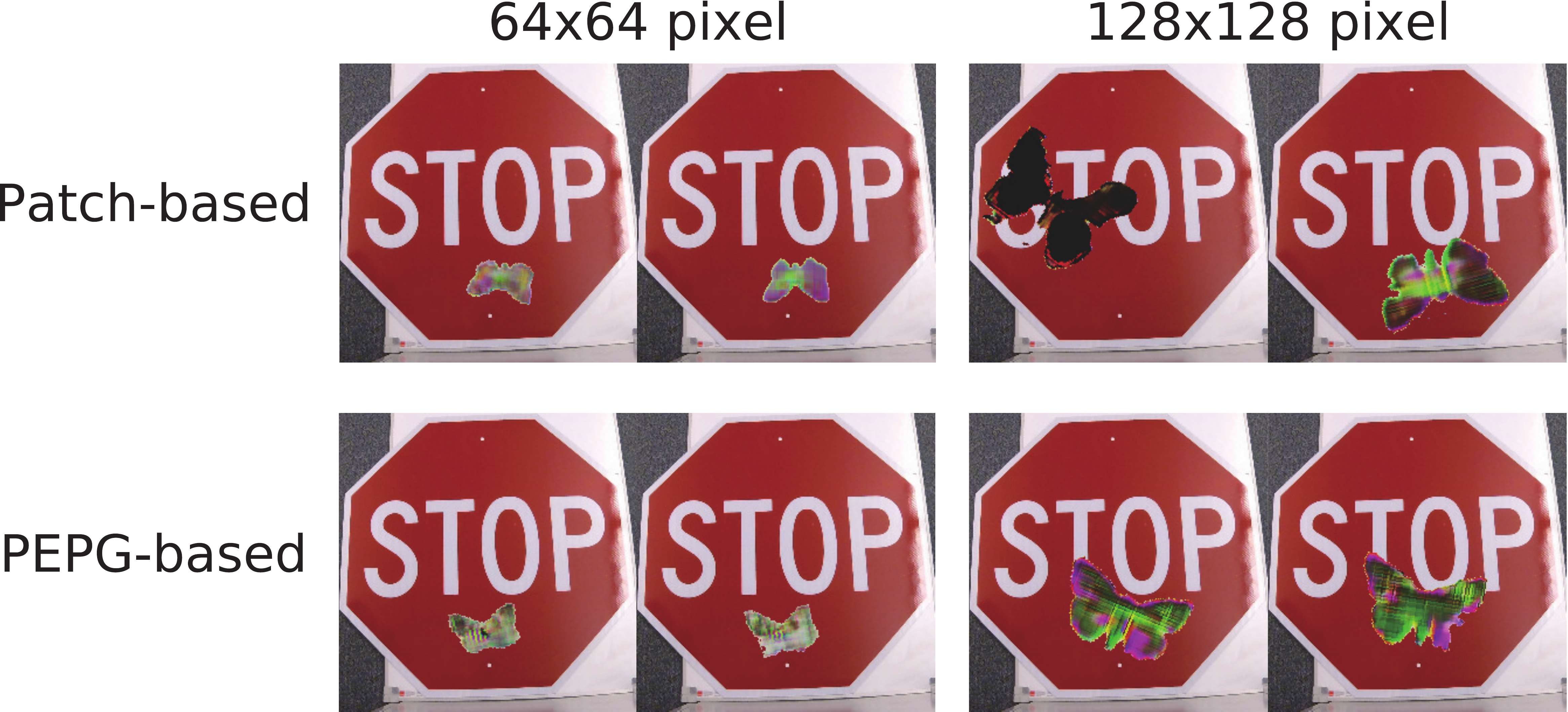}
    \caption{Examples of adversarial examples that made the road sign classifier recognize them as ``Speed Limit 80.''}
    \label{fig:road-ae}
\end{figure}

We first explored the feasibility of the proposed methods with a road sign classifier trained on the German Traffic Sign Recognition Benchmark \cite{DBLP:journals/nn/StallkampSSI12}, as done by \citealp{DBLP:conf/cvpr/EykholtEF0RXPKS18}.
We used their convolutional neural network-based classifier for the target model\footnote{\url{https://github.com/evtimovi/robust_physical_perturbations}}, which was reported to show 95.7\% accuracy.
For the input image, we used the same images of stop signs measuring $256 \times 256$ pixels and tried to make them be recognized as Speed Limit 80 in the same manner as done by \citealp{DBLP:conf/cvpr/EykholtEF0RXPKS18}.
Regarding each combination of the perturbation size and generation approach, we examined whether we could obtain adversarial examples in a given number of iterations.

The obtained adversarial examples are presented in \figref{fig:road-ae}.
The result indicates that the success of the generation highly depends on the size of the perturbation; that is, we could not generate an adversarial example after 20,000 iterations when the perturbation size was limited to $32 \times 32$.
This corresponds to the result from \citealp{DBLP:journals/corr/abs-1712-09665}: the larger the adversarial patch becomes, the higher the success rate became.

By comparing the patch-based and PEPG-based methods, we found that the PEPG-based method took much less time.
For example, in the case of $128 \times 128$ pixels, the PEPG-based method took about 6 minutes and 753 iterations to generate 100 adversarial examples, whereas the patch-based method took about an hour and 5,340 iterations.
This difference suggests that optimization of the location and rotation of the perturbation helps achieve success by finding the sensitive region of the input image.

\subsection{ImageNet Classifier}
\label{sec:results-imagenet}

\begin{figure}[t]
    \centering
    \includegraphics[width=0.48\textwidth]{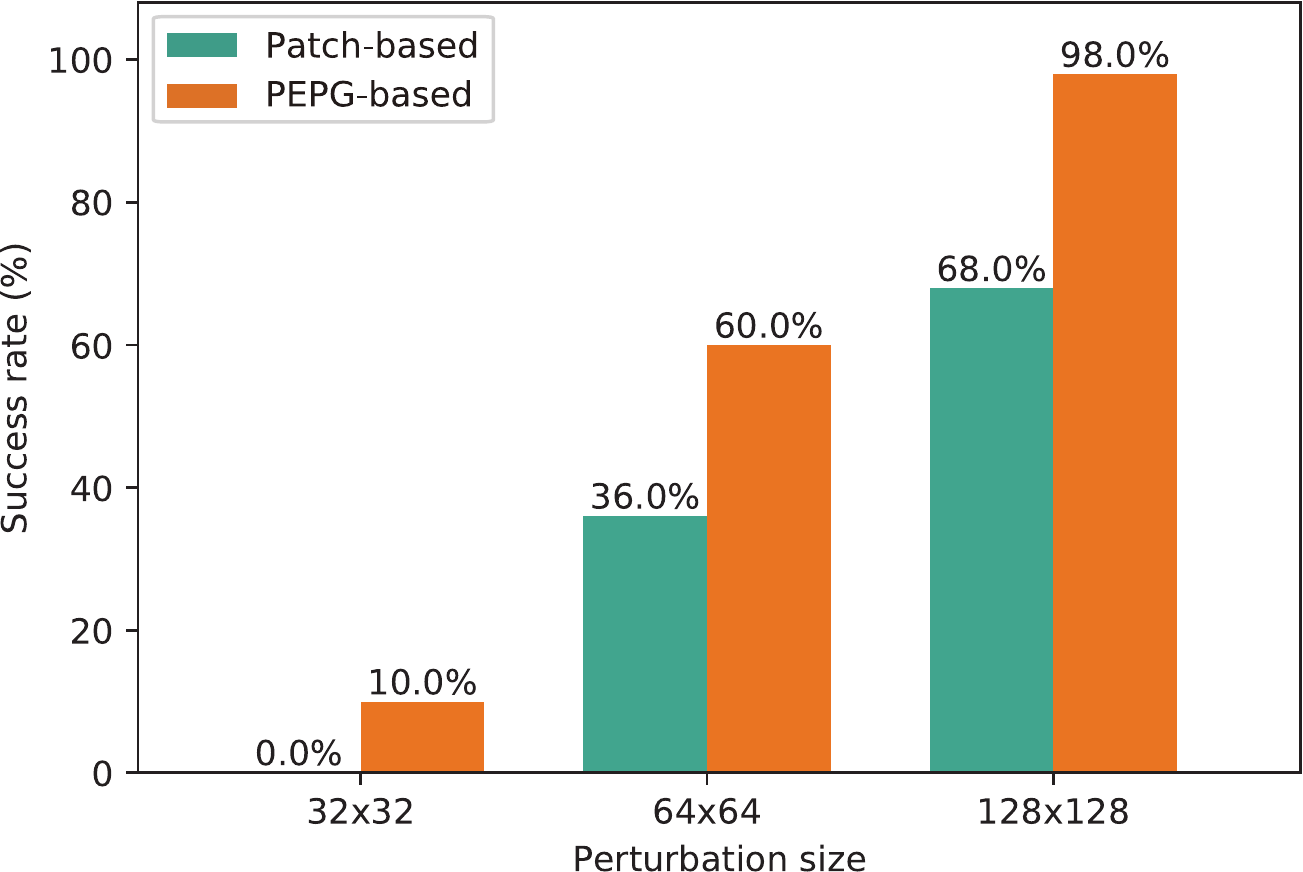}
    \caption{Success rate for each combination of perturbation size and generation approach against the ImageNet classifier.}
    \label{fig:imagenet-success}
\end{figure}

\begin{figure}[t]
    \centering
    \includegraphics[width=0.48\textwidth]{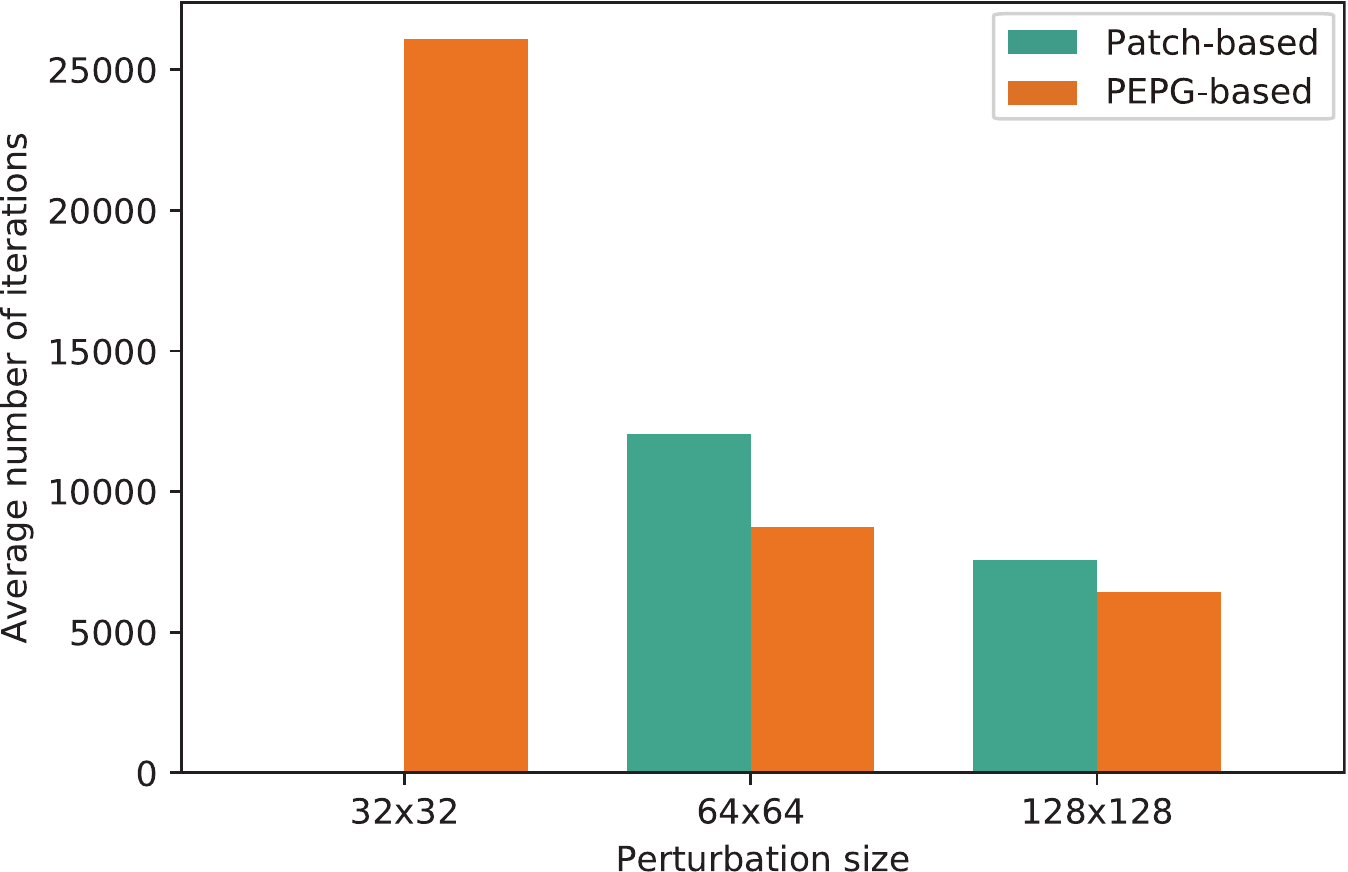}
    \caption{Average number of iterations required to generate 100 successful adversarial examples.}
    \label{fig:imagenet-iteration}
\end{figure}

We then evaluated the proposed methods using an Inception-V3 classifier \cite{DBLP:conf/cvpr/SzegedyVISW16} pretrained on ImageNet, which is distributed by TensorFlow\footnote{\url{https://github.com/tensorflow/models/tree/master/research/slim}}.
For the combination of the input image and target class, we used the same tasks as the NIPS '17 competition on adversarial examples \cite{DBLP:journals/corr/abs-1804-00097}.
Here, because the proposed methods involve a training process for each combination and require some time, we selected the first 50 tasks.

In the same manner described in \secref{sec:results-sign}, we examined whether we could obtain 100 adversarial examples in 50,000 iterations for each combination of perturbation size and generation approach.
Then, we compared the success rate over 50 tasks and the average number of iterations in successful cases.

\begin{figure}[t]
    \begin{center}
        \includegraphics[width=0.47\textwidth]{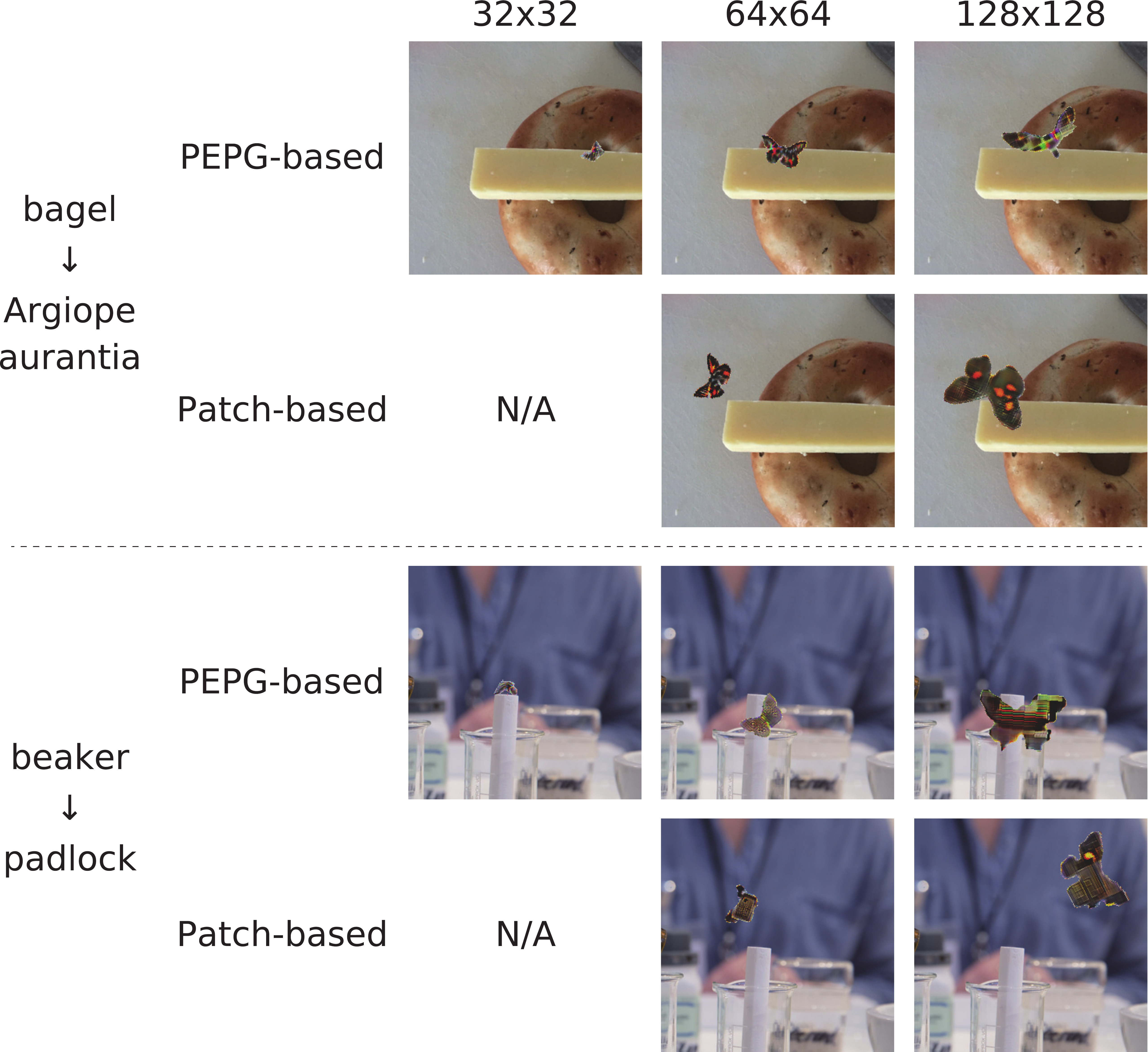}
        \caption{Examples of adversarial examples generated against the ImageNet classifier.}
        \label{fig:imagenet-ae}
    \end{center}
\end{figure}

The results are shown in \figref{fig:imagenet-success} and \figref{fig:imagenet-iteration}.
Some images of the obtained adversarial examples are also shown in \figref{fig:imagenet-ae}.
We confirmed that the larger size of the perturbation helps the generation of adversarial examples; that is, it increases the success rate and decreases the required number of iterations.

We also confirmed that the optimization of the perturbation location by the PEPG algorithm helps the generation.
In particular, the PEPG-based method succeeded in generating adversarial examples with perturbations of only $32 \times 32$ pixels in five tasks, whereas the patch-based method completely failed.
In addition, in the case of $128 \times 128$ pixels, the PEPG-based method succeeded in generating 100 adversarial examples for 49 tasks.
We also note that, in the remaining (50th) task, our method could not generate 100 examples within 50,000 iterations but obtained 91 adversarial examples.
From these points, the proposed approach of optimizing the location is shown to be effective for increasing the success rate.

\subsection{Analysis}
\label{sec:results-analysis}

In this section, we analyze characteristics of the proposed method based on the above results and additional investigations.
We first examine the robustness of the perturbations obtained by the patch-based method.
Then, we examine the effectiveness of the PEPG algorithm for the successful generation.

\subsubsection{Robustness of the Patch-based Perturbations}
\label{sec:results-analysis-patch}

\begin{figure}[t]
    \centering
    \includegraphics[width=0.48\textwidth]{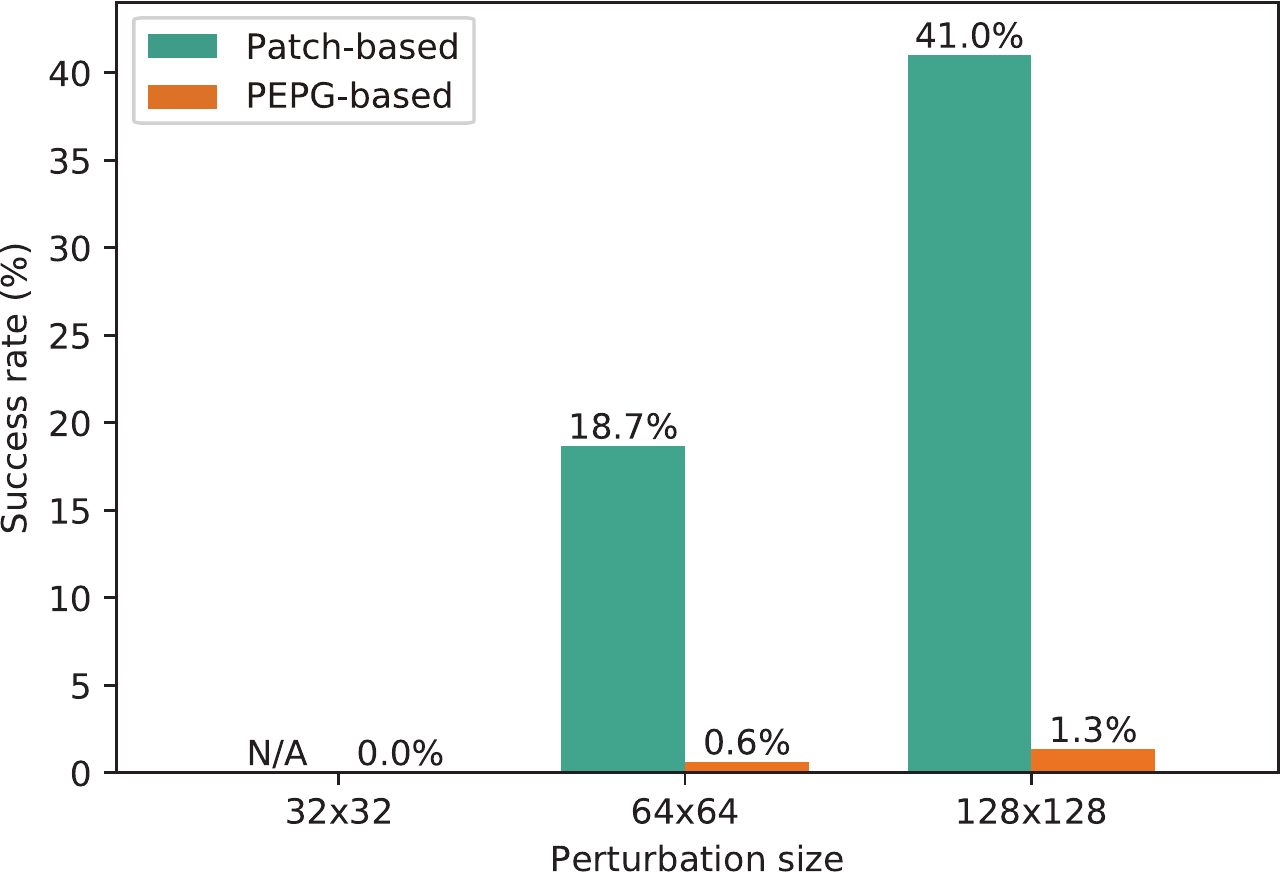}
    \caption{Ratio of samples crafted by random relocation of the perturbations and classified into the target label.}
    \label{fig:robustness}
\end{figure}

As discussed in \secref{sec:method-patch}, the patch-based method generates perturbations that are robust against change in location.
Thus, it is expected that we can produce new adversarial examples by relocating the perturbations in a fashion similar to that described in \citealp{DBLP:journals/corr/abs-1712-09665}.
We tested the idea by crafting 10 samples from each adversarial example obtained as described in \secref{sec:results-imagenet} with a random relocation and verifying their classification result.

\begin{figure}[t]
    \centering
    \includegraphics[width=0.45\textwidth]{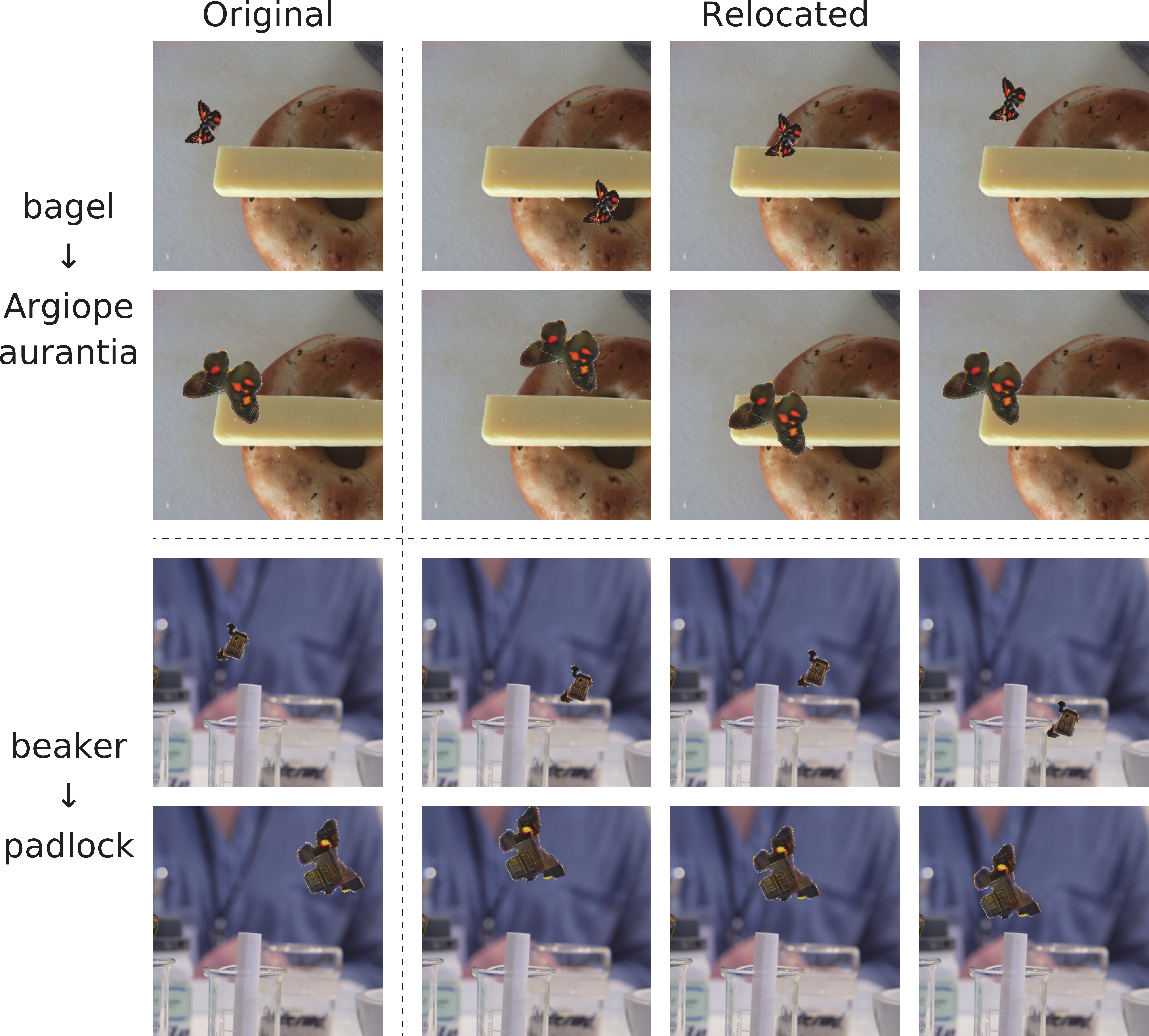}
    \caption{Examples of successful samples crafted from adversarial examples obtained by the patch-based method.}
    \label{fig:robustness-patch}
\end{figure}

\begin{figure}[t]
    \centering
    \includegraphics[width=0.28\textwidth]{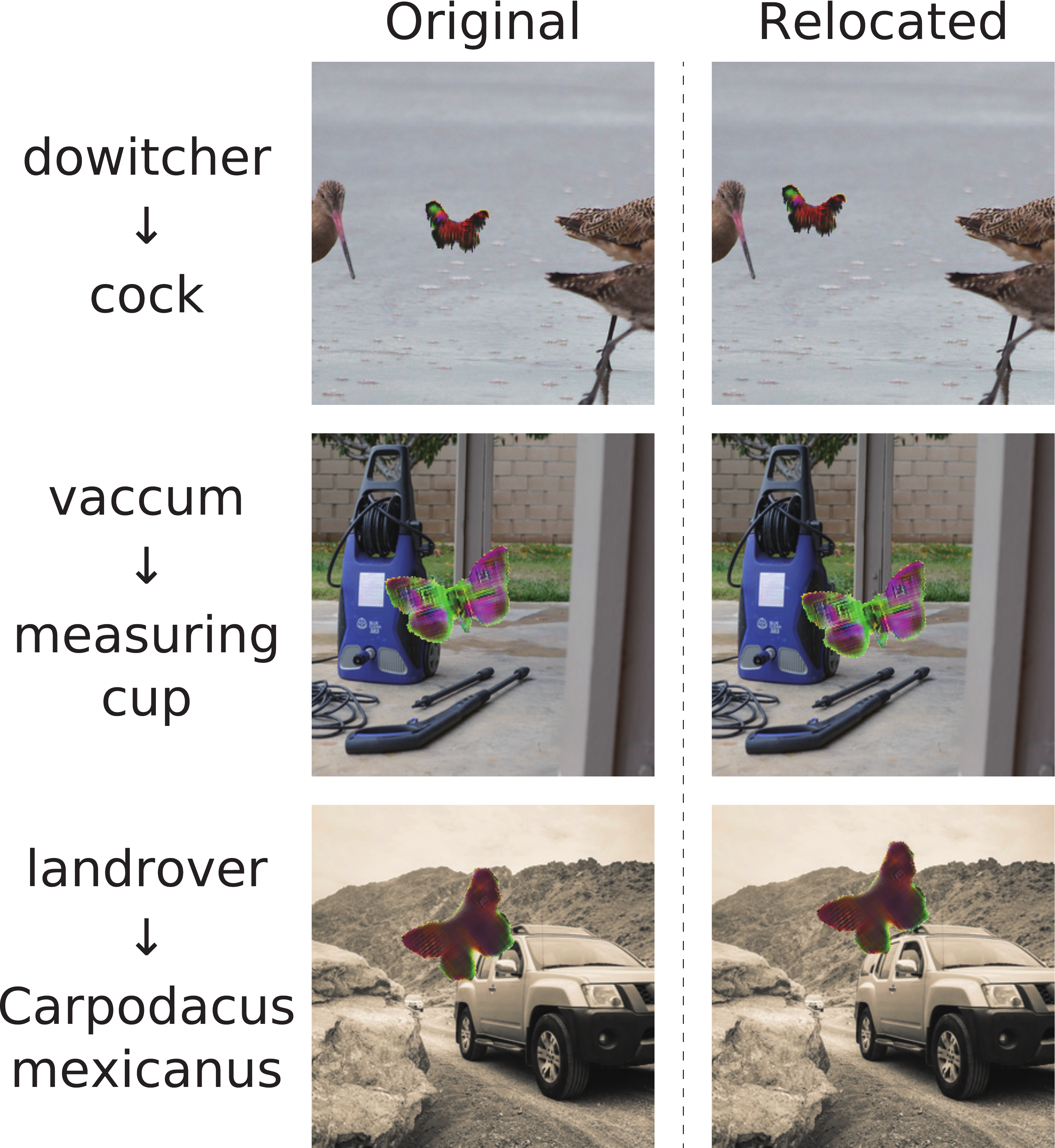}
    \caption{Examples of successful samples crafted from adversarial examples obtained by the PEPG-based method.}
    \label{fig:robustness-pepg}
\end{figure}

\figref{fig:robustness} shows the success rate for adversarial example multiplication, that is, the ratio of the crafted samples that are classified to the same target label as the original adversarial example.
These results suggest that the perturbations obtained by the patch-based method are more robust than the results for the PEPG-based method.
In particular, we found that successful samples among the adversarial examples obtained by the patch-based method (\figref{fig:robustness-patch}) have perturbations in various locations,
whereas the samples obtained from the PEPG-based method (\figref{fig:robustness-pepg}) exhibit a limited number of successful cases where the perturbation location is similar to that in the original adversarial examples.
We note that a success rate is roughly comparable to the adversarial patch \cite{DBLP:journals/corr/abs-1712-09665},
which showed the success rate of about 20\% when modifying 3\% of the pixels in the image\footnote{The perturbation of $64 \times 64$ pixels accounts for only 4.6\% of the image of $299 \times 299$ pixels.},
although the proposed method uses a GAN to generate perturbations instead of optimizing them directly.

\subsubsection{Effectiveness of the PEPG Algorithm}
\label{sec:results-analysis-pepg}

We confirmed that the perturbations obtained by the PEPG-based method show limited robustness regarding the relocation.
Conversely, it implies that the PEPG algorithm successfully finds the limited area where the perturbation functions as an adversarial example.
For example, in \figref{fig:imagenet-ae}, we found that all adversarial examples obtained by the PEPG-based method have perturbations in similar locations in the same input image.

\begin{figure}[t]
    \centering
    \includegraphics[width=0.37\textwidth]{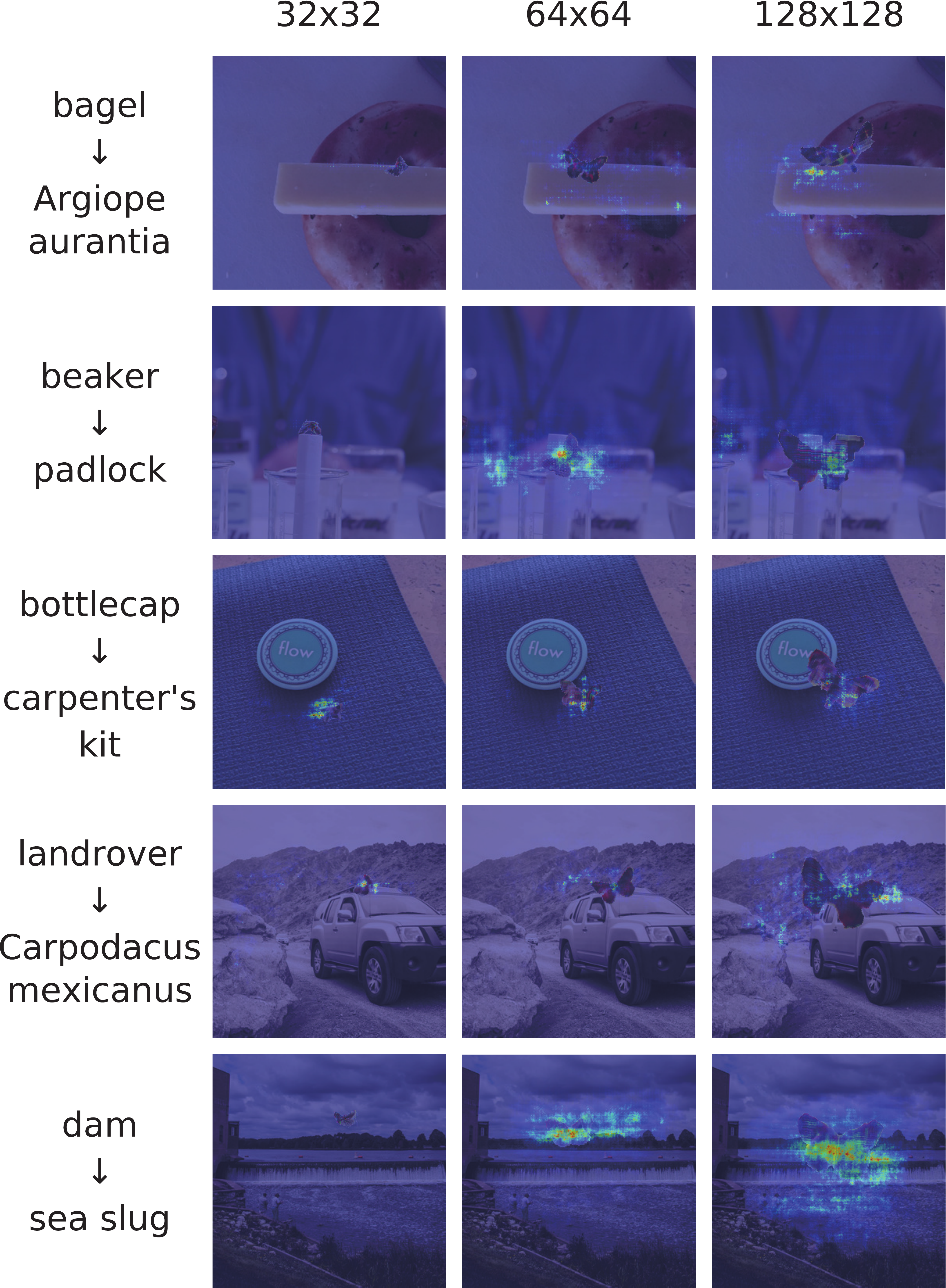}
    \caption{Confidence of the target label obtained when the center of the perturbation is shifted to each pixel.}
    \label{fig:effectiveness-location}
\end{figure}

Thus, we examined how the classification results are changed when we move the perturbations to a different location.
The heatmaps of \figref{fig:effectiveness-location} show the confidence that the image obtained by shifting the center of the perturbation to each pixel is classified as the target label.
The results confirm that the PEPG algorithm successfully finds the limited location where the perturbation functions as an adversarial example, especially when the perturbation is small.

\begin{figure}[t]
    \centering
    \includegraphics[width=0.46\textwidth]{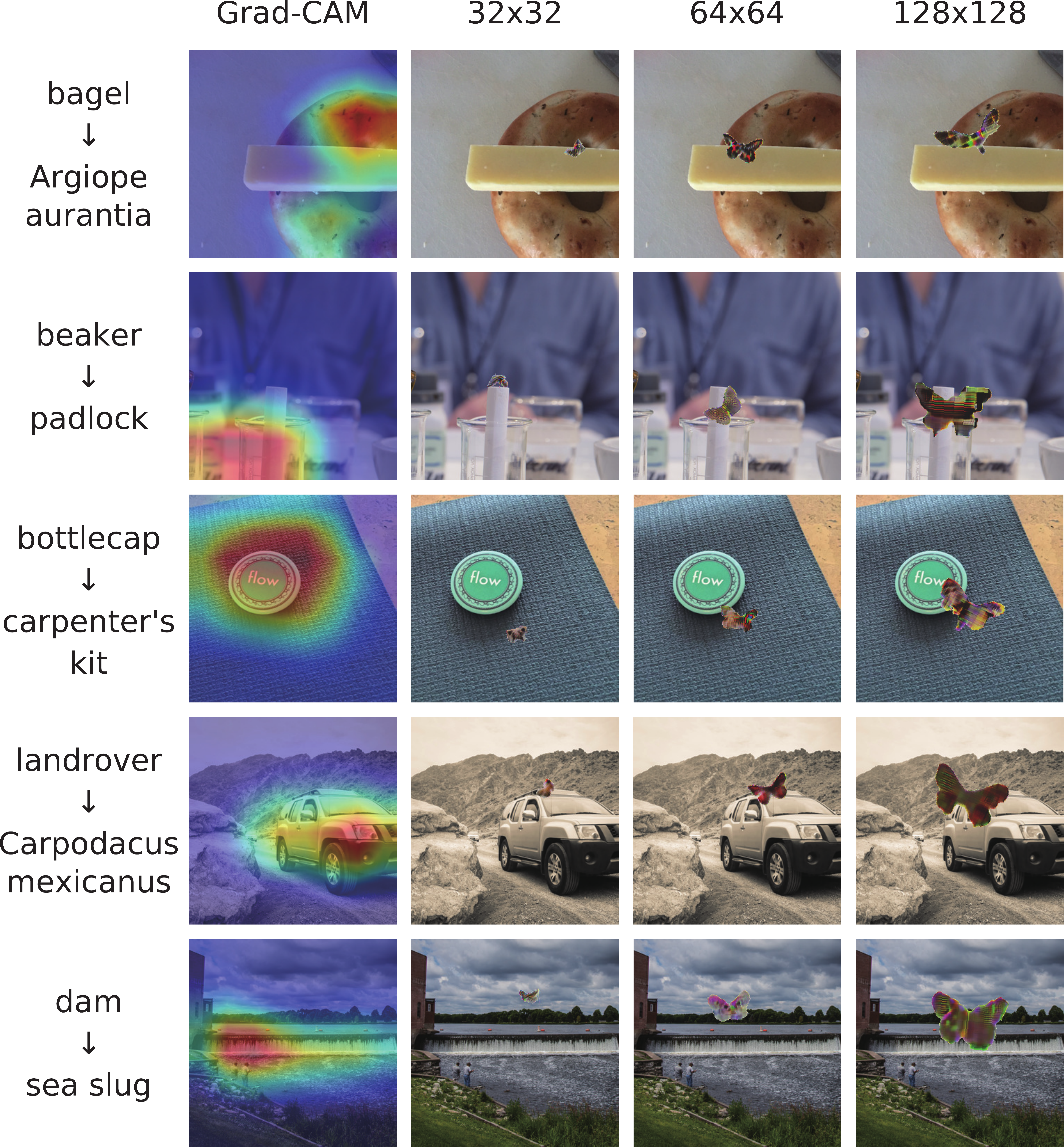}
    \caption{Comparison between the activation maps of the input image and the location of the perturbation optimized by the PEPG-based method.}
    \label{fig:effectiveness-cam}
\end{figure}

In addition, we compared these adversarial examples with the activation heatmaps of the corresponding input image obtained by Grad-CAM \cite{DBLP:conf/iccv/SelvarajuCDVPB17}, which shows the class-discriminative regions.
The results are shown in \figref{fig:effectiveness-cam}.
We found that the perturbations were placed near the most discriminative regions in the input images, regardless of the size of the perturbations.
These analyses support the effectiveness of the PEPG-based method for optimizing the perturbation placement, which would contribute to its higher success rate compared to the patch-based method, as presented in \secref{sec:results-imagenet}.

In summary, our analysis suggests the existence of some trade-off between the robustness and the success rate when applying the patch-based and PEPG-based method, as discussed in \secref{sec:method}.
In other words, we can choose an appropriate method from them to fit different attack situations.

\section{Application: Audio Adversarial Example}
\label{sec:audio}

In this paper, we presented a new approach for generating adversarial examples by making perturbations imitate a specific object.
Up to this point, we discussed attacks that targeting image classification models, but this idea can be applied to other domains.
For example, if we can generate an adversarial example that fools speech recognition models but sounds like the chirping of small birds,
we would be able to control personal assistants based on speech interactions (e.g., Siri or Google Assistant) without being noticed by humans.
To investigate the potential of the proposed approach, we tested whether we can generate such audio adversarial examples using GAN in the same manner as used for images (\secref{sec:method}).

As with image adversarial examples, there are many proposed methods for generating audio adversarial examples \cite{DBLP:journals/corr/abs-1801-00554,DBLP:conf/sp/Carlini018,DBLP:conf/ijcai/YakuraS19,DBLP:conf/ndss/SchonherrKZHK19}.
For example, \citealp{DBLP:journals/corr/abs-1801-00554} generated adversarial examples against a speech command classifier and attained a success rate of 87\%.
\citealp{DBLP:conf/sp/Carlini018} proposed a generation method that targets DeepSpeech \cite{DBLP:journals/corr/HannunCCCDEPSSCN14} by extending their previous work \cite{DBLP:conf/sp/Carlini017} from image adversarial examples.
\citealp{DBLP:conf/ijcai/YakuraS19} realized a physical attack against DeepSpeech by simulating noise and reverberation in the physical world during audio generation.
With respect to perceptibility by humans, \citealp{DBLP:conf/ndss/SchonherrKZHK19} optimized less noticeable perturbations by exploiting a psychoacoustic model.
However, none of them tried to manipulate the content of the perturbation to match the attack scenario, such as camouflaging the perturbation with environmental noise like birds chirping.

\subsection{Settings}
\label{sec:audio-settings}

For the target model, we used the same speech command classifier \cite{DBLP:conf/interspeech/SainathP15} as \citealp{DBLP:journals/corr/abs-1801-00554},
which is distributed by TensorFlow\footnote{\url{https://www.tensorflow.org/tutorials/sequences/audio_recognition}}.
We employed the architecture WaveGAN \cite{DBLP:conf/iclr/DonahueMP19}, which is based on WGAN-GP \cite{DBLP:conf/nips/GulrajaniAADC17},
to generate perturbations\footnote{The source code is available at \url{https://github.com/hiromu/adversarial_examples_with_bugs}.}.

For the reference audio, we used the VB100 Bird Dataset \cite{DBLP:conf/dicta/GeMSWLRC16} to make perturbations that sounded like chirping birds.
The generated perturbations were added to one of two audio clips from the Speech Commands Dataset \cite{DBLP:journals/corr/abs-1804-03209}, which says ``yes'' or ``no.''
Then, we examined whether the obtained adversarial examples were classified as the target label; in this case, we chose ``stop'' among 12 labels of the model output that included ''yes'' and ''no.''

\subsection{Results}
\label{sec:audio-results}

We succeeded in generating adversarial examples that were classified as ``stop,'' although they sound like someone saying ``yes'' or ``no'' with chirping birds in a background.
Some of the obtained adversarial examples are available at \url{https://yumetaro.info/projects/bugs-ae/}.

We also conducted a listening experiment with 25 human participants using Amazon Mechanical Turk in a manner similar to that in \citealp{DBLP:journals/corr/abs-1801-00554}.
First, we presented two obtained adversarial examples made from the clips saying ``yes'' and ''no'' and asked the participants to write down the words they heard.
Then, we asked them to write down anything abnormal that they detected.
The responses are summarized as follows:
\begin{itemize}
    \setlength{\leftskip}{8pt}
    \setlength{\itemsep}{4pt}
    \setlength{\parskip}{0pt}
    \setlength{\parsep}{0pt}
    \item For the transcription tasks:
    \begin{itemize}
        \setlength{\leftskip}{10pt}
        \setlength{\itemsep}{2pt}
        \setlength{\parskip}{0pt}
        \setlength{\parsep}{0pt}
        \item All participants transcribed the contents of the original clips.
        \item No participant identified the target word.
    \end{itemize}
    \item In the collected comments:
    \begin{itemize}
        \setlength{\leftskip}{10pt}
        \setlength{\itemsep}{2pt}
        \setlength{\parskip}{0pt}
        \setlength{\parsep}{0pt}
        \item Six participants noted the existence of birds chirping in the background.
        \item No participant felt suspicious about the clips.
    \end{itemize}
\end{itemize}
These responses suggest that the proposed approach for making perturbations mimic unnoticeable objects also works in the audio domain and has very low perception for humans.

\section{Limitations and Future Directions}
\label{sec:limitations}

While our results confirmed the effectiveness of the new approach of generating adversarial examples by making perturbations imitate a specific object or signal, there are still some limitations.
In particular, though we experimentally investigated the effectiveness of the PEPG-based method in \secref{sec:results-analysis}, its theoretical background is still unclear.
One possible clue would be a transition of the reward of the PEPG algorithm, considering that literature on reinforcement learning often use it to examine the training process in detail.

In this respect, further investigations are needed to clarify the prospects of this new idea of applying the PEPG algorithm to the generation of adversarial examples.
For example, though we fixed the size of the perturbations in \secref{sec:results}, it would be possible also to optimize the size if we can design an appropriate penalty term.
For the audio adversarial examples, the start timing of synthesizing the perturbation can be optimized so as to find the best position to hide the birds chirping.

The discussion of the attack scenarios is also an important research direction.
As discussed in \secref{sec:introduction}, the proposed approach can increase the magnitude of the perturbation without being noticed by humans.
Given that many defense methods have been defeated with a relatively small magnitude of the perturbation \cite{DBLP:conf/ccs/Carlini017}, this approach potentially opens up further attack possibilities,
which we must discuss to ensure the safety of socio-technical systems based on machine learning.

\section{Conclusion}
\label{sec:conclusion}

In this paper, we proposed a systematic approach for generating natural adversarial examples by making perturbations imitate specific objects or signals, such as bugs images.
We presented its feasibility for attacking image classifiers by leveraging GAN to generate perturbations that imitated the reference images and fooled the target model.
We also confirmed that the optimization of the perturbation location using the PEPG algorithm led to the successful generation of adversarial examples.
Furthermore, we experimentally showed that the proposed approach could be extended to the audio domain, such as generating a perturbation that sounds like the chirping of birds.
Our results provide a new direction for creating natural adversarial examples by manipulating the content of the perturbation instead of trying to make the perturbation imperceptible by limiting its magnitude.

\section*{Acknowledgments}

This work was supported by KAKENHI 19H04164 and 18H04099.

\bibliography{AAAI-YakuraH.3716}

\begin{thebibliography}{}

\bibitem[\protect\citeauthoryear{Alzantot, Balaji, and
  Srivastava}{2017}]{DBLP:journals/corr/abs-1801-00554}
Alzantot, M.; Balaji, B.; and Srivastava, M.~B.
\newblock 2017.
\newblock Did you hear that? adversarial examples against automatic speech
  recognition.
\newblock In {\em Proceedings of the 2017 {NIPS} Workshop on Machine
  Deception},  1--6.
\newblock San Diego, CA: {NIPS} Foundation.

\bibitem[\protect\citeauthoryear{Brown \bgroup et al\mbox.\egroup
  }{2017}]{DBLP:journals/corr/abs-1712-09665}
Brown, T.~B.; Man{\'{e}}, D.; Roy, A.; Abadi, M.; and Gilmer, J.
\newblock 2017.
\newblock Adversarial patch.
\newblock In {\em Proceedings of the 2017 {NIPS} Workshop on Machine Learning
  and Computer Security},  1--5.
\newblock San Diego, CA: {NIPS} Foundation.

\bibitem[\protect\citeauthoryear{Carlini and
  Wagner}{2017a}]{DBLP:conf/ccs/Carlini017}
Carlini, N., and Wagner, D.~A.
\newblock 2017a.
\newblock Adversarial examples are not easily detected: Bypassing ten detection
  methods.
\newblock In {\em Proceedings of the 10th {ACM} Workshop on Artificial
  Intelligence and Security},  3--14.
\newblock New York, NY: {ACM}.

\bibitem[\protect\citeauthoryear{Carlini and
  Wagner}{2017b}]{DBLP:conf/sp/Carlini017}
Carlini, N., and Wagner, D.~A.
\newblock 2017b.
\newblock Towards evaluating the robustness of neural networks.
\newblock In {\em Proceedings of the 38th {IEEE} Symposium on Security and
  Privacy},  39--57.
\newblock Washington, DC: {IEEE} Computer Society.

\bibitem[\protect\citeauthoryear{Carlini and
  Wagner}{2018}]{DBLP:conf/sp/Carlini018}
Carlini, N., and Wagner, D.~A.
\newblock 2018.
\newblock Audio adversarial examples: Targeted attacks on speech-to-text.
\newblock In {\em Proceedings of the 1st {IEEE} Deep Learning and Security
  Workshop},  1--7.
\newblock Washington, DC: {IEEE} Computer Society.

\bibitem[\protect\citeauthoryear{Chen \bgroup et al\mbox.\egroup
  }{2018}]{DBLP:conf/pkdd/ChenCMC18}
Chen, S.; Cornelius, C.; Martin, J.; and Chau, D. H.~P.
\newblock 2018.
\newblock Shapeshifter: Robust physical adversarial attack on faster {R-CNN}
  object detector.
\newblock In {\em Proceedings of the 2018 European Conference on Machine
  Learning and Principles and Practice of Knowledge Discovery in Databases},
  volume 11051,  52--68.
\newblock Cham, Switzerland: Springer.

\bibitem[\protect\citeauthoryear{Donahue, McAuley, and
  Puckette}{2019}]{DBLP:conf/iclr/DonahueMP19}
Donahue, C.; McAuley, J.~J.; and Puckette, M.~S.
\newblock 2019.
\newblock Adversarial audio synthesis.
\newblock In {\em Proceedings of the 7th International Conference on Learning
  Representations},  1--16.
\newblock La Jolla, CA: {ICLR}.

\bibitem[\protect\citeauthoryear{Eykholt \bgroup et al\mbox.\egroup
  }{2018}]{DBLP:conf/cvpr/EykholtEF0RXPKS18}
Eykholt, K.; Evtimov, I.; Fernandes, E.; Li, B.; Rahmati, A.; Xiao, C.;
  Prakash, A.; Kohno, T.; and Song, D.
\newblock 2018.
\newblock Robust physical-world attacks on deep learning visual classification.
\newblock In {\em Proceedings of the 31st {IEEE/CVF} Conference on Computer
  Vision and Pattern Recognition},  1625--1634.
\newblock Washington, DC: {IEEE} Computer Society.

\bibitem[\protect\citeauthoryear{Ge \bgroup et al\mbox.\egroup
  }{2016}]{DBLP:conf/dicta/GeMSWLRC16}
Ge, Z.; McCool, C.; Sanderson, C.; Wang, P.; Liu, L.; Reid, I.~D.; and Corke,
  P.~I.
\newblock 2016.
\newblock Exploiting temporal information for dcnn-based fine-grained object
  classification.
\newblock In {\em Proceedings of the 2016 International Conference on Digital
  Image Computing: Techniques and Applications},  1--6.
\newblock New York, NY: {IEEE}.

\bibitem[\protect\citeauthoryear{Goodfellow \bgroup et al\mbox.\egroup
  }{2014}]{DBLP:conf/nips/GoodfellowPMXWOCB14}
Goodfellow, I.~J.; Pouget{-}Abadie, J.; Mirza, M.; Xu, B.; Warde{-}Farley, D.;
  Ozair, S.; Courville, A.~C.; and Bengio, Y.
\newblock 2014.
\newblock Generative adversarial nets.
\newblock In {\em Proceedings of the 28th Annual Conference on Neural
  Information Processing Systems},  2672--2680.
\newblock San Diego, CA: {NIPS} Foundation.

\bibitem[\protect\citeauthoryear{Goodfellow, Shlens, and
  Szegedy}{2015}]{DBLP:journals/corr/GoodfellowSS14}
Goodfellow, I.~J.; Shlens, J.; and Szegedy, C.
\newblock 2015.
\newblock Explaining and harnessing adversarial examples.
\newblock In {\em Proceedings of the 3rd International Conference on Learning
  Representations},  1--11.
\newblock La Jolla, CA: {ICLR}.

\bibitem[\protect\citeauthoryear{Grosse \bgroup et al\mbox.\egroup
  }{2017}]{DBLP:conf/esorics/GrossePMBM17}
Grosse, K.; Papernot, N.; Manoharan, P.; Backes, M.; and McDaniel, P.~D.
\newblock 2017.
\newblock Adversarial examples for malware detection.
\newblock In {\em Proceedings of the 22nd European Symposium on Research in
  Computer Security}, volume 10493,  62--79.
\newblock Cham, Switzerland: Springer.

\bibitem[\protect\citeauthoryear{Gulrajani \bgroup et al\mbox.\egroup
  }{2017}]{DBLP:conf/nips/GulrajaniAADC17}
Gulrajani, I.; Ahmed, F.; Arjovsky, M.; Dumoulin, V.; and Courville, A.~C.
\newblock 2017.
\newblock Improved training of wasserstein gans.
\newblock In {\em Proceedings of the 31st Annual Conference on Neural
  Information Processing Systems},  5769--5779.
\newblock San Diego, CA: {NIPS} Foundation.

\bibitem[\protect\citeauthoryear{Hannun \bgroup et al\mbox.\egroup
  }{2014}]{DBLP:journals/corr/HannunCCCDEPSSCN14}
Hannun, A.~Y.; Case, C.; Casper, J.; Catanzaro, B.; Diamos, G.; Elsen, E.;
  Prenger, R.; Satheesh, S.; Sengupta, S.; Coates, A.; and Ng, A.~Y.
\newblock 2014.
\newblock Deep speech: Scaling up end-to-end speech recognition.
\newblock {\em arXiv} 1412.5567:1--12.

\bibitem[\protect\citeauthoryear{Jaderberg \bgroup et al\mbox.\egroup
  }{2015}]{DBLP:conf/nips/JaderbergSZK15}
Jaderberg, M.; Simonyan, K.; Zisserman, A.; and Kavukcuoglu, K.
\newblock 2015.
\newblock Spatial transformer networks.
\newblock In {\em Proceedings of the 29th Annual Conference on Neural
  Information Processing Systems},  2017--2025.
\newblock San Diego, CA: {NIPS} Foundation.

\bibitem[\protect\citeauthoryear{Jia and Liang}{2017}]{DBLP:conf/emnlp/JiaL17}
Jia, R., and Liang, P.
\newblock 2017.
\newblock Adversarial examples for evaluating reading comprehension systems.
\newblock In {\em Proceedings of the 2017 Conference on Empirical Methods in
  Natural Language Processing},  2021--2031.
\newblock Stroudsburg, PA: {ACL}.

\bibitem[\protect\citeauthoryear{Kurakin \bgroup et al\mbox.\egroup
  }{2018}]{DBLP:journals/corr/abs-1804-00097}
Kurakin, A.; Goodfellow, I.~J.; Bengio, S.; Dong, Y.; Liao, F.; Liang, M.;
  Pang, T.; Zhu, J.; Hu, X.; Xie, C.; Wang, J.; Zhang, Z.; Ren, Z.; Yuille,
  A.~L.; Huang, S.; Zhao, Y.; Zhao, Y.; Han, Z.; Long, J.; Berdibekov, Y.;
  Akiba, T.; Tokui, S.; and Abe, M.
\newblock 2018.
\newblock Adversarial attacks and defences competition.
\newblock In {\em The NIPS '17 Competition: Building Intelligent Systems}.
  Cham, Switzerland: Springer.
\newblock  195--231.

\bibitem[\protect\citeauthoryear{LeCun, Bengio, and
  Hinton}{2015}]{DBLP:journals/nature/LeCunBH15}
LeCun, Y.; Bengio, Y.; and Hinton, G.~E.
\newblock 2015.
\newblock Deep learning.
\newblock {\em Nature} 521(7553):436--444.

\bibitem[\protect\citeauthoryear{Moosavi{-}Dezfooli, Fawzi, and
  Frossard}{2016}]{DBLP:conf/cvpr/Moosavi-Dezfooli16}
Moosavi{-}Dezfooli, S.; Fawzi, A.; and Frossard, P.
\newblock 2016.
\newblock Deepfool: {A} simple and accurate method to fool deep neural
  networks.
\newblock In {\em Proceedings of the 29th {IEEE} Conference on Computer Vision
  and Pattern Recognition},  2574--2582.
\newblock Washington, DC: {IEEE} Computer Society.

\bibitem[\protect\citeauthoryear{Ng, Yang, and
  Davis}{2015}]{DBLP:conf/cvpr/NgYD15}
Ng, J.~Y.; Yang, F.; and Davis, L.~S.
\newblock 2015.
\newblock Exploiting local features from deep networks for image retrieval.
\newblock In {\em Proceedings of the 2015 {CVPR} Workshop on Deep Vision},
  53--61.
\newblock Washington, DC: {IEEE} Computer Society.

\bibitem[\protect\citeauthoryear{Rodner \bgroup et al\mbox.\egroup
  }{2015}]{DBLP:journals/corr/RodnerSBPWD15}
Rodner, E.; Simon, M.; Brehm, G.; Pietsch, S.; W{\"{a}}gele, J.; and Denzler,
  J.
\newblock 2015.
\newblock Fine-grained recognition datasets for biodiversity analysis.
\newblock In {\em Proceedings of the 2015 {CVPR} Workshop on Fine-grained
  Visual Classification},  1--4.
\newblock Washington, DC: {IEEE} Computer Society.

\bibitem[\protect\citeauthoryear{Sainath and
  Parada}{2015}]{DBLP:conf/interspeech/SainathP15}
Sainath, T.~N., and Parada, C.
\newblock 2015.
\newblock Convolutional neural networks for small-footprint keyword spotting.
\newblock In {\em Proceedings of the 16th Annual Conference of the
  International Speech Communication Association},  1478--1482.
\newblock Baixas, France: {ISCA}.

\bibitem[\protect\citeauthoryear{Sch{\"{o}}nherr \bgroup et al\mbox.\egroup
  }{2019}]{DBLP:conf/ndss/SchonherrKZHK19}
Sch{\"{o}}nherr, L.; Kohls, K.; Zeiler, S.; Holz, T.; and Kolossa, D.
\newblock 2019.
\newblock Adversarial attacks against automatic speech recognition systems via
  psychoacoustic hiding.
\newblock In {\em Proceedings of the 26th Annual Network and Distributed System
  Security Symposium},  1--15.
\newblock Reston, VA: The Internet Society.

\bibitem[\protect\citeauthoryear{Sehnke \bgroup et al\mbox.\egroup
  }{2010}]{DBLP:journals/nn/SehnkeORGPS10}
Sehnke, F.; Osendorfer, C.; R{\"{u}}ckstie{\ss}, T.; Graves, A.; Peters, J.;
  and Schmidhuber, J.
\newblock 2010.
\newblock Parameter-exploring policy gradients.
\newblock {\em Neural Networks} 23(4):551--559.

\bibitem[\protect\citeauthoryear{Selvaraju \bgroup et al\mbox.\egroup
  }{2017}]{DBLP:conf/iccv/SelvarajuCDVPB17}
Selvaraju, R.~R.; Cogswell, M.; Das, A.; Vedantam, R.; Parikh, D.; and Batra,
  D.
\newblock 2017.
\newblock Grad-cam: Visual explanations from deep networks via gradient-based
  localization.
\newblock In {\em Proceedings of the 16th {IEEE} International Conference on
  Computer Vision},  618--626.
\newblock Washington, DC: {IEEE} Computer Society.

\bibitem[\protect\citeauthoryear{Sharif \bgroup et al\mbox.\egroup
  }{2016}]{DBLP:conf/ccs/SharifBBR16}
Sharif, M.; Bhagavatula, S.; Bauer, L.; and Reiter, M.~K.
\newblock 2016.
\newblock Accessorize to a crime: Real and stealthy attacks on state-of-the-art
  face recognition.
\newblock In {\em Proceedings of the 23rd {ACM} {SIGSAC} Conference on Computer
  and Communications Security},  1528--1540.
\newblock New York, NY: {ACM}.

\bibitem[\protect\citeauthoryear{Stallkamp \bgroup et al\mbox.\egroup
  }{2012}]{DBLP:journals/nn/StallkampSSI12}
Stallkamp, J.; Schlipsing, M.; Salmen, J.; and Igel, C.
\newblock 2012.
\newblock Man vs. computer: Benchmarking machine learning algorithms for
  traffic sign recognition.
\newblock {\em Neural Networks} 32:323--332.

\bibitem[\protect\citeauthoryear{Su, Vargas, and
  Sakurai}{2017}]{DBLP:journals/corr/abs-1710-08864}
Su, J.; Vargas, D.~V.; and Sakurai, K.
\newblock 2017.
\newblock One pixel attack for fooling deep neural networks.
\newblock {\em arXiv} 1710.08864:1--15.

\bibitem[\protect\citeauthoryear{Szegedy \bgroup et al\mbox.\egroup
  }{2014}]{DBLP:journals/corr/SzegedyZSBEGF13}
Szegedy, C.; Zaremba, W.; Sutskever, I.; Bruna, J.; Erhan, D.; Goodfellow,
  I.~J.; and Fergus, R.
\newblock 2014.
\newblock Intriguing properties of neural networks.
\newblock In {\em Proceedings of the 2nd International Conference on Learning
  Representations},  1--10.
\newblock La Jolla, CA: {ICLR}.

\bibitem[\protect\citeauthoryear{Szegedy \bgroup et al\mbox.\egroup
  }{2016}]{DBLP:conf/cvpr/SzegedyVISW16}
Szegedy, C.; Vanhoucke, V.; Ioffe, S.; Shlens, J.; and Wojna, Z.
\newblock 2016.
\newblock Rethinking the inception architecture for computer vision.
\newblock In {\em Proceedings of the 29th {IEEE} Conference on Computer Vision
  and Pattern Recognition},  2818--2826.
\newblock Washington, DC: {IEEE} Computer Society.

\bibitem[\protect\citeauthoryear{Warden}{2018}]{DBLP:journals/corr/abs-1804-03209}
Warden, P.
\newblock 2018.
\newblock Speech commands: {A} dataset for limited-vocabulary speech
  recognition.
\newblock {\em arXiv} 1804.03209:1--11.

\bibitem[\protect\citeauthoryear{Xiao \bgroup et al\mbox.\egroup
  }{2018}]{DBLP:conf/ijcai/XiaoLZHLS18}
Xiao, C.; Li, B.; Zhu, J.; He, W.; Liu, M.; and Song, D.
\newblock 2018.
\newblock Generating adversarial examples with adversarial networks.
\newblock In {\em Proceedings of the 27th International Joint Conference on
  Artificial Intelligence},  3905--3911.
\newblock {IJCAI}.

\bibitem[\protect\citeauthoryear{Yakura and
  Sakuma}{2019}]{DBLP:conf/ijcai/YakuraS19}
Yakura, H., and Sakuma, J.
\newblock 2019.
\newblock Robust audio adversarial example for a physical attack.
\newblock In {\em Proceedings of the 28th International Joint Conference on
  Artificial Intelligence},  5334--5341.
\newblock {IJCAI}.

\bibitem[\protect\citeauthoryear{Zhao, Dua, and
  Singh}{2018}]{DBLP:conf/iclr/ZhaoDS18}
Zhao, Z.; Dua, D.; and Singh, S.
\newblock 2018.
\newblock Generating natural adversarial examples.
\newblock In {\em Proceedings of the 6th International Conference on Learning
  Representations},  1--15.
\newblock La Jolla, CA: {ICLR}.

\end{thebibliography}
\bibliographystyle{aaai}

\end{document}